\documentclass[11pt]{article}
\usepackage{geometry}
\usepackage{coling2020}
\usepackage{times}
\usepackage{url}
\usepackage{color}
\usepackage{latexsym}
\usepackage{microtype}
\usepackage{amsmath}
\usepackage{graphicx}
\usepackage{subcaption}

\colingfinalcopy % Uncomment this line for all SemEval submissions

\newcommand{\ourmodel}{SChME}
\newcommand{\coderepo}{\url{https://github.com/mgruppi/schme}}

\title{\ourmodel~at SemEval-2020 Task 1:  A Model Ensemble for Detecting Lexical Semantic Change}

\author{Maur\'{i}cio Gruppi, Sibel Adal{\i} \\
  Rensselaer Polytechnic Institute \\
  Troy, NY, USA \\
  {\tt \{gouvem, adalis\}@rpi.edu} \\\And
  Pin-Yu Chen \\
  IBM Research \\
  Yorktown Heights, NY, USA\\
  {\tt pin-yu.chen@ibm.com}
  }

\date{}

\begin{document}
\maketitle
\begin{abstract}
  This paper describes \ourmodel~(Semantic Change Detection with Model Ensemble), a method used in SemEval-2020 Task 1 on unsupervised detection of lexical semantic change. \ourmodel~uses a model ensemble combining signals of distributional models (word embeddings) and word frequency models where each model casts a vote indicating the probability that a word suffered semantic change according to that feature. More specifically, we combine cosine distance of word vectors combined with a neighborhood-based metric we named Mapped Neighborhood Distance (MAP), and a word frequency differential metric as input signals to our model. Additionally, we explore alignment-based methods to investigate the importance of the landmarks used in this process. Our results show evidence that the number of landmarks used for alignment has a direct impact on the predictive performance of the model. Moreover, we show that languages that suffer less semantic change tend to benefit from using a large number of landmarks, whereas languages with more semantic change benefit from a more careful choice of landmark number for alignment.
\end{abstract}

\section{Introduction}

The problem of detecting Lexical Semantic Change (LSC) consists of measuring and identifying change in word sense across time, such as in the study of language evolution, or across domains, such as determining discrepancies in word usage over specific communities \cite{schlechtweg-etal-2019-wind}. 
One of the greatest challenges of this problem is the difficulty of assessing and evaluating models and results, as well as the limited amount of annotated data \cite{schlechtweg2020simulating}.
For that reason, the vast majority of the related work in the literature pursue this problem from an unsupervised perspective, that is, detecting semantic change without having prior knowledge of ``truth''. The importance of such task is manifold: to humans, it can be a powerful tool for studying language change and its cultural implications; to machines, it can be used to improve language models in downstream tasks such as unsupervised word translation, and fine-tuning of word embeddings \cite{joulin2018loss,bojanowski2019updating}. In this task, the goal is to develop a method for unsupervised detection of lexical semantic change over time by comparing across two corpora from different time periods in four languages: English, German, Latin, and Swedish \cite{schlechtweg2020semeval}. Particularly, we are required to solve two sub-tasks: binary classification of semantic change (\textbf{Subtask 1}), and semantic change ranking (\textbf{Subtask 2}).

\blfootnote{This work is licensed under a Creative Commons Attribution 4.0 International License. License details: \url{http://creativecommons.org/licenses/by/4.0/}.}

% \begin{itemize}
%     \item {\textbf{Subtask 1}: binary classification - detecting whether or not the occurrence of semantic change for a word.}
%     \item {\textbf{Subtask 2}: ranking - grade the degree in which word sense-frequencies have changed.}
% \end{itemize}

There are many ways in which a word may change. Specifically, a word $w$ may change sense because it has been completely replaced by a synonym $w_s$ (lexical replacement), or because it gains a new meaning, in which case word $w$ may keep or lose its previous meaning across time and domain \cite{kutuzov-etal-2018-diachronic}. 
Each type of change has its unique characteristics and may require different approaches in order to be detected.
In this paper we describe a novel model ensemble method based on different features (signals) that we can extract from the text using distribution models (skip-gram word embeddings) and word frequency. Our model is primarily based on features extracted from independently trained Word2Vec embeddings aligned with orthogonal procrustes \cite{schonemann1966generalized}, such as cosine distance, but also introduces two novel measures based on second-order distances and word frequency. Based on the distribution of each feature, we predict the probability that a word has suffered change through an anomaly detection approach. The final decision is made by soft voting (averaging) all the probabilities. For binary classification (Subtask 1) a threshold is applied to the final vote, for ranking (Subtask 2), the output from the soft voting is used as the ranking prediction.

Our results show that second order methods and different combinations outperform the frequently used cosine distance in some subtasks and languages. Furthermore, we illustrate that the methods are sensitive to the degree of change in the language. It is possible to improve performance of these methods by aligning two embeddings of the same language from different time slices on a subset of words instead of all words. This opens a new avenue of research on finding optimal words for alignment.
The code for the model can be obtained at \coderepo.

\section{Related Work}

Most methods for detecting semantic change are based on the distributional property of word semantics. The general idea is to compute contextual information of word $w$ in each time or domain, and apply a measure of difference or distance between the observed contexts of $w$. Some of the first methods for detecting semantic change compute context information using a co-occurrence matrix within a pre-defined window of size $L$ \cite{sagi2009semantic,cook2010automatically}. This means that, for a vocabulary of size $n$, one computes a $n \times n$ matrix $M$ where $M_{i,j}$ is the frequency in which word $i$ and $j$ co-occur within a window of $L$ words. This often yields a highly sparse matrix $M$, which is typically reduced in dimensionality by techniques such as Singular Value Decomposition (SVD). Once the matrices are computed, the contextual difference is computed by the cosine distance between the vectors.

Distributed word vector representations such as the ones obtained by the skip-gram with negative sampling (SGNS) \cite{mikolov2013distributed} are forms of learning distributional information without the need for computing sparse co-occurrence matrices. 
A work by \newcite{hamilton2016diachronic} presents a method for detecting semantic change using SGNS word embeddings learned from each corpora and aligned with orthogonal procrustes. 
The semantic change is, again, computed by the cosine distance between vectors in each time/domain. 
In another study \cite{hamilton2016cultural}, the authors introduce a measure of semantic change based on how the neighborhood of a word changes named Local Neighborhood Change based on the number of words in common. 
% This metric is computed by the cosine distance between second-order vectors of similarities for word vectors of a word in each domain.

To eliminate the need for alignment, several authors have proposed dynamic word embeddings techniques, which jointly learn distributional word representations using the assumption that words are connected across time \cite{bamler2017dynamic,rudolph2018dynamic,yao2018dynamic}. 
The main assumption in such methods is that word changes are considerably small between adjacent time stamps $t_1$ and $t_2$, i.e. words evolve smoothly, thus word representations should be close between these periods.
We argue that the assumption that all words in $t_1$ and $t_2$ should be smoothly connected through time does not always hold.
This is because the corpora are aggregated over several years/decades/centuries, thus the semantic change may be drastic, and more similar to a cross-domain scenario than a diachronic one.
We illustrate this by the corpora in this task and the use of a subset of landmarks for alignment that has not been investigated in the literature.  

\section{Model Overview and Data}

The data provided in this task consists of two corpora for each language, each corpus corresponding to different time periods $t_1$ and $t_2$, as well as a list of target words for which we have to predict binary class and rank with respect to magnitude of the semantic change between $t_1$ and $t_2$. The corpora used for each language are summarized in Table \ref{tab:data}.

\begin{table}[ht]
    \centering
    \begin{tabular}{c|c|c|c}
         \textbf{Language}   &  \textbf{Corpora}                          & $t_1$   & $t_2$   \\ \hline
         \textbf{English}    &  CCOHA \cite{alatrash2020clean}   & 1810-1860   & 1960-2010   \\
         \textbf{German}     &  DTA + BZ + ND \cite{schlechtweg2020semeval} & 1800-1900   & 1946-1990   \\
         \textbf{Latin}      &  LatinISE \cite{mcgillivray2013tools}  & -200 - 0   & 0-2000  \\
         \textbf{Swedish}    &  KubHist \cite{borin-etal-2012-korp} & 1790-1830   & 1895-1903
    \end{tabular}
    \caption{Data provided for the task. In addition to the corpora, a set of target words is given, for which we need to generate outputs in substasks 1 and 2.}
    \label{tab:data}
\end{table}

\subsection{Word Representations}
Most of our features are based on the alignment of word embeddings. Thus, the first step of our system is to train a Word2Vec model on corpora $C_1$ and $C_2$ for each language, let $W_1$ and $W_2$ denote the resulting word embeddings, respectively. Since $W_1$ and $W_2$ are learned independently, we cannot directly compare their vectors. Hence, similarly to \newcite{hamilton2016diachronic}, we apply orthogonal procrustes (OP) \cite{schonemann1966generalized} to align the word embeddings of the corpora. Given matrices $A$ and $B$, the objective of OP is to learn an orthogonal transformation matrix $Q$ that minimizes the sum of squared distances $\left\lVert AQ - B \right\rVert_2$. Because $Q$ is orthogonal, the transformation $AQ$ is only subject to rotation and reflection, which preserves the relationships between the word vectors in $A$. We learn the transformation matrix $Q$ from the alignment of $W_1$ and $W_2$, updating $W_1 \leftarrow W_1Q$. Now the word vectors in $W_1$ can be directly compared to $W_2$.
In the following sections, we'll discuss the distance metrics used by the model to measure semantic change.

\subsection{Distance Measures}

\paragraph{Cosine Distance (COS).}
One of the most used metric for comparing word vectors is the cosine distance. The cosine distance between two vectors in a single source indicates how closely distributed the words are. In the semantic change scenario, we compute the cosine distance for word $w$ as $d_{cos} = 1 - cos(v_1, v_2)$, where $v_1$ and $v_2$ are the word vectors of $w$ in $W_1$ and $W_2$, respectively. Ideally, a small value of $d_{cos}$ would imply that the contexts for $w$ is similar in both corpora $C_1$ and $C_2$.

\paragraph{Mapped Neighborhood Change (MAP).}
This measure looks at how a word moves away from its neighborhood across both corpora. To that end, we compute a second-order cosine distance vector $s_1(v_1, \mathcal{N}_1)$ between $v_1$ and its $k$ nearest neighbors in $W_1$, which we'll denote as the set $\mathcal{N}_1$. Then we compute another second-order vector $s_2(v_1, \mathcal{N}_1)$ using $v_1$ but looking for corresponding vectors of each word in $\mathcal{N}_1$ in the space of the second corpus $W_2$. The mapped neighborhood change is then computed as the cosine distance $ d_{map}(v_1) = d_{cos}(s_1(v_1, \mathcal{N}_1), s_2(v_1, \mathcal{N}_1)) $. Although this method uses second-order distances like the Local Neighborhood Change (LNC) \cite{hamilton2016cultural}, it differs from it by computing the distances between the aligned input embeddings, while LNC only computes such distances for vectors within a single embedding matrix.

% Note that this operation is made possible due to the alignment (mapping) between $W_1$ and $W_2$. Moreover, this operation is asymmetric, meaning $d_{map}(v_1(w)) \neq d_{map}(v_2(w))$ because the $k$ nearest neighbors of $v_1(w)$ and $v_2(w)$ are not necessarily the same. 

\paragraph{Frequency Differential (FREQ).}
Let $f_1$ and $f_2$ be the relative frequencies of word $w$ in $C_1$ and $C_2$. We define the frequency differential for $w$ as $ f(w) = \frac{f_1 - f_2}{f_1 + f_2} $. Positive values indicate increase while negative values indicate decrease in frequency across the corpora. We argue that a steep increase in frequency may indicate indicate change more strongly than frequency decrease, which may happen due to a word becoming less popular or being replaced by another word without losing its original sense. This assumption is only viable because we know that $C_1$ always happens earlier in time than $C_2$.

\subsection{Model Ensemble}
We compute the aforementioned features on all words in the intersection of the vocabularies of $C_1$ and $C_2$, we use the observed feature distributions to determine potentially changed words. Let $\mathbf{X}_i$ denote the random variable associated with the distribution of feature $i$. We work under the assumption that small values of $\mathbf{X}_i$ denote little or no semantic change to a word. Moreover, unlikely high values of $\mathbf{X}_i$ indicate a high chance that the word suffered change according to metric $i$. We define small and large values with respect to all the computed values in the distribution. For instance, if the cosine distance computed for a word is large when compared to the cosine distances of the other words, it is likely that the word has changed. Therefore, we define the probability of change for a word whose feature value is $x_i$ as $ P_i(x_i) = Pr(\mathbf{X}_i \leq x_i)$.

Thus, $P_i$ is the cumulative distribution function (CDF) of $\mathbf{X}_i$, describing how unlikely high $x_i$ is according to the distribution of $\mathbf{X}_i$. We aggregate the probability output of each feature $P_i(x_i)$ by applying soft voting to each feature's prediction. The final prediction for a feature vector $\mathbf{x} = (x_1, x_2, ..., x_k)$ is $ P(\mathbf{x}) = \frac{1}{k}\sum_{1}^{k}P_i(x_i)$. For classification, a threshold is a applied to $P(\mathbf{x})$ in order to determine the class. For ranking, the score $P(\mathbf{x})$ is used directly.

% \begin{figure}[ht]
%     \centering
%     \includegraphics[width=0.48\textwidth]{model.pdf}
%     \caption{Overview of the ensemble model with soft voting. Votes from each feature model $\mathbf{M}_i$ are aggregated by averaging and to output function $f(x)$. For binary classification, $f(x)$ applies a threshold to $ P(\mathbf{x}) $ to decide the class. For ranking, we use the identify function $f(x)=x$.}
%     \label{fig:model}
% \end{figure}

\section{Evaluation}

We conduct all the experiments on the data provided for SemEval-2020 Task 1 for all four languages. Given that most of the corpora have been pre-processed with lemmatization and tokenization, our pre-processing consists of removing words whose count is less than 10, and tokenizing words at spaces. In this section we present the experiments and results for the model submitted to the task, as well as additional analysis of the model parameters.

We begin by learning the distributional representations of words in each corpora using Gensim's \cite{rehurek_lrec} implementation of Word2Vec . The parameters for Word2Vec are: vector size $d=300$, window $L=10$, negative samples $ng=5$, and minimum word count $min\_wc = 10$. Next, we align the learned word vectors via OP using the intersecting vocabulary as landmarks. Then, we compute the distance metrics and their distributions so that we can get the vote $Pr(\mathbf{X}_i \leq x_i)$. Finally, we apply the model ensemble to different feature configurations to predict a final score. For classification, we apply a threshold $t$ to the model output $P(\mathbf{x})$, such that the predicted class is $y = 1$ if $P(\mathbf{x}) > t$, and $y=0$ otherwise. For ranking, the final score $P(\mathbf{x})$ is used.

Since there was no validation data during the evaluation phase, our submissions included multiple feature and threshold settings. The feature configurations are combinations of the cosine distance (COS), mapped neighborhood distance (MAP), and frequency differential (FREQ). The applied threshold levels are $\{0.5, 0.75, 0.9\}$.
Our team (RPI-Trust) ranked 4th place in Subtask 1 with a score of 0.660, and 6th place in Subtask 2 with a score of 0.427 in the evaluation phase.

\begin{table*}[t]
    \centering
    \begin{tabular}{l|rrrrr}
        Feature & English & German & Latin & Swedish & Decay (\%) \\ \hline
         COS &          0.622   &   \textbf{0.75}   &     0.45    &   \textbf{0.806}  & 0.09   \\
         MAP &          0.595   &   0.604           &     \textbf{0.575}   &   0.677  & 0.14 \\
         COS+FREQ &     0.595   &   \textbf{0.75}   &     0.525   &   0.742       & 0.09 \\
         COS+MAP+FREQ &  \textbf{0.649}  &   0.729  &     0.45    &   0.742       & 0.10 \\
         Maj. Class &   (0)0.568    &   (0)0.646   &   (1)0.650 & (0)0.742 & - \\  \hline
    \end{tabular}
    \caption{Classification accuracy for different feature configurations at a threshold $t=0.75$. Majority class (Maj. Class) is a baseline classifier that outputs the most common class for each language (classes 0 or 1). Column decay indicates the accuracy deviation from the best performance for each feature model across languages. A smaller decay means the method performs close to optimal in all languages.}
    \label{tab:classification}
\end{table*}

\begin{table*}[t]
    \centering
    \begin{tabular}{l|rrrrr}
        Feature & English & German & Latin & Swedish & Decay (\%)  \\ \hline
         COS &  \textbf{0.231}      & \textbf{0.547}     &   0.413      &  0.228    & 0.09 \\
         MAP &        0.05          & 0.504     &  0.388                &  0.200  & 0.32 \\
         COS+FREQ &   0.26          & 0.407     &  \textbf{0.455}       &  -0.009  & 0.32  \\
         COS+MAP+FREQ & 0.203       & 0.433     & 0.424                 &   \textbf{0.268}  & 0.12    \\ \hline
    \end{tabular}
    \caption{Ranking performance (Spearman's $\rho$) for each feature model. Column decay indicates the Spearman's $\rho$ deviation from the best performing method in each language. Decay is defined in Table~\ref{tab:classification}.}
    \label{tab:ranking}
\end{table*}

\section{Post-Evaluation}

We evaluate our model on the provided test data in the post-evaluation phase. First, we fix a threshold of $t=0.75$, then we use different feature combinations to evaluate the performance on each language. Classification results, seen in Table \ref{tab:classification}, show that there is no single best feature configuration for all languages. This may happen because each language evolved differently between $t_1$ and $t_2$, and having each feature model being able to capture different types of change. For example, many events in between $t_1$ and $t_2$ for the English corpora may have contributed to the evolution of the language, such as the Second Industrial Revolution, and the World Wars. Technological development introduced several new concepts such as (air) \emph{plane} and (record) \emph{player} which were unheard of in $t_1$, the detection of such change relies on signals that can indicate a completely new use of a word while potentially keeping its previous senses. The results for the ranking task are shown in Table \ref{tab:ranking}. Notice that the best feature configurations for classification are not necessarily the best for ranking. MAP performs best for Latin which might be due to potential big semantic shift in this language which is better captured by incorporating neighborhood information. As seen in the \emph{decay} column, COS and COS+MAP+FREQ (used in our submission) are the overall best performing methods across the two tasks.

% \subsection{Threshold Sensitivity}

% We conduct an experiment in the post-evaluation phase to analyze the threshold sensitivity of each feature configuration for the classification task. To that end, we analyze the model's accuracy when varying the threshold in the interval $(0, 1)$. The classification threshold 

\subsection{Landmarks Are Important}

\begin{figure}[ht]
    \centering
    \begin{subfigure}{0.4\textwidth}
    \centering
    \includegraphics[width=1.0\textwidth]{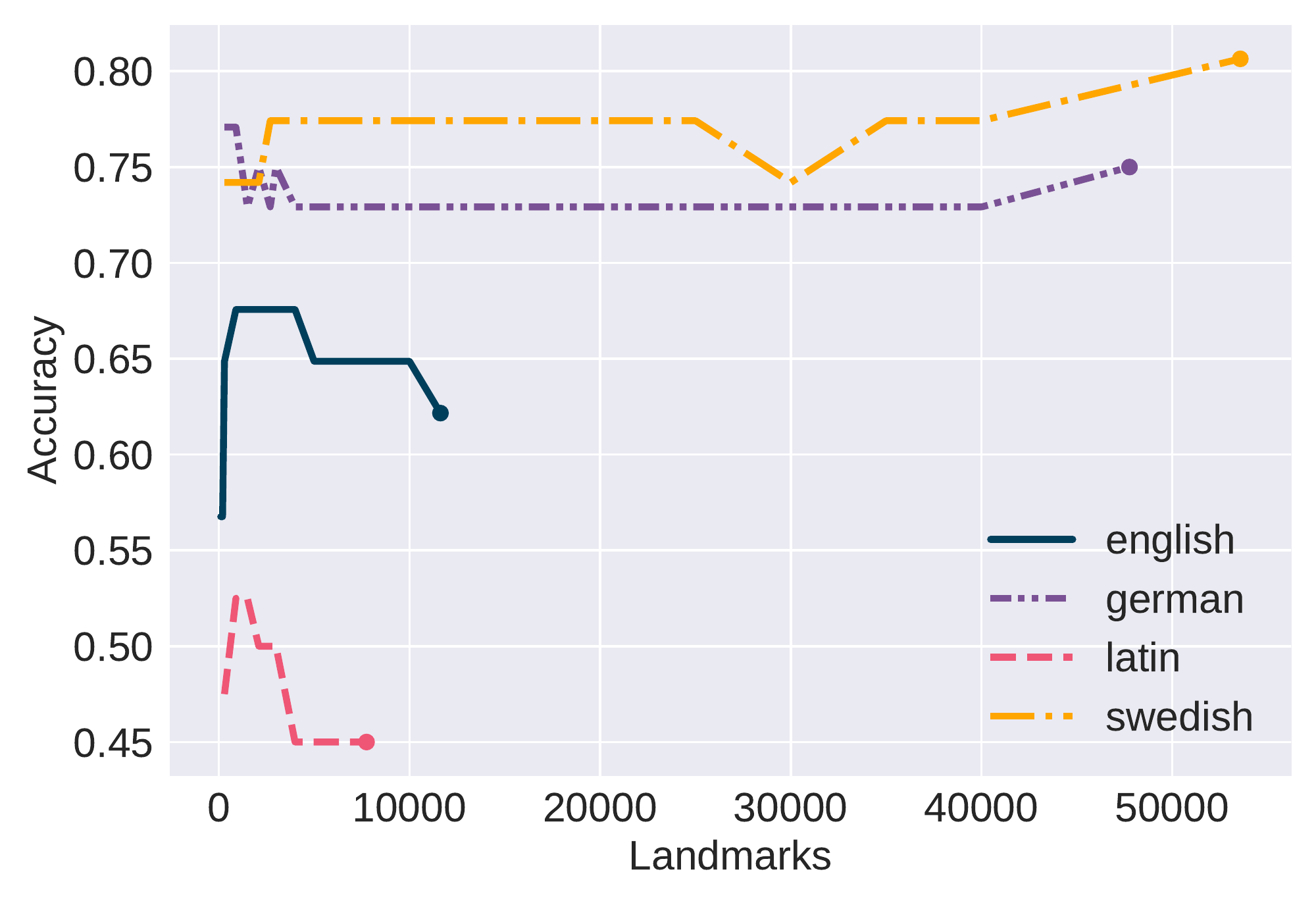}
    \caption{}
    \end{subfigure}\qquad
    \begin{subfigure}{0.4\textwidth}
    \centering
    \includegraphics[width=1.0\textwidth]{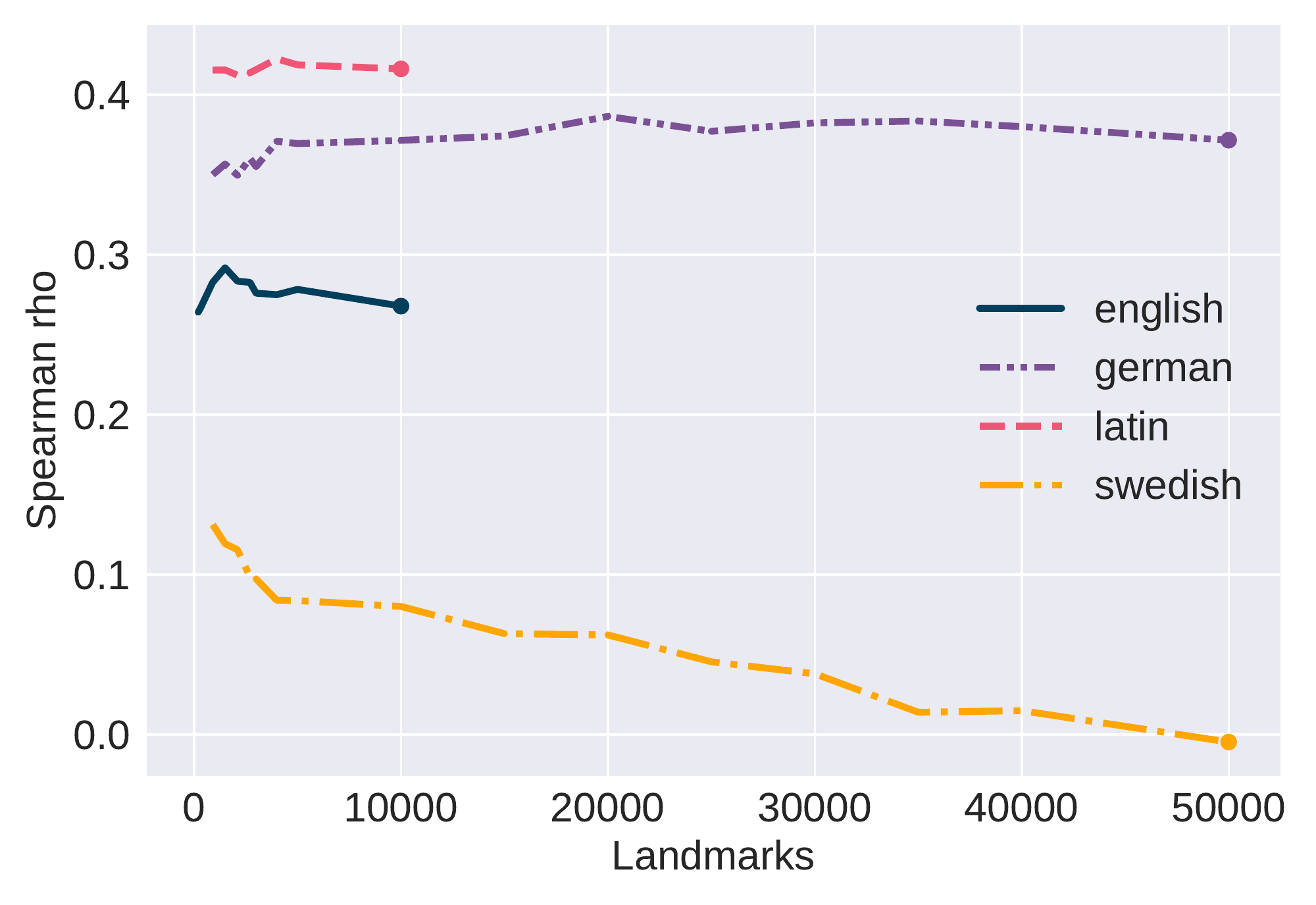}
    \caption{}
    \end{subfigure}
    
    \caption{(a) Accuracy in Subtask 1 using different numbers of landmark words for each language. Notice how German and Swedish do not show a decrease in accuracy despite the large number of landmarks used, whereas English and Latin have optimal performance at some point before the maximum; (b) Ranking performance according to number of landmarks shows a different trend from that of the binary classification with Swedish decreasing in performance as the number of landmarks grow.}
    \label{fig:landmarks_cls}
    
\end{figure}

When executing procrustes alignment, one must choose which and how many words to align on. Since alignment seeks to enforce short distances between landmark words, we hypothesize that this method may mask some of the semantic shift involving the landmark words. To test this, we analyze the effect of the number of landmark words over the model predictions by executing procrustes alignment at using the top $n$ most frequent landmark words with $n \in [300, N]$, where $N$ is the size of the intersecting vocabulary, keeping a classifier threshold fixed at $t=0.75$. Figure \ref{fig:landmarks_cls} shows the results for all four languages.

These results present evidence to our argument: using more landmark words in the alignment procedure favors German and Swedish that likely have less semantic shift compared to Latin and English. Notice that both corpora present class imbalance leaning towards unchanged words, and show increased accuracy as the number of landmarks increase. On the other hand, the same is not true for English, which has more balanced classes, nor for Latin which is unbalanced towards changed words. In both these languages, the classification accuracy peaks at some $n < N$ and then decreases, thus showing that using all possible words as landmarks may decrease the accuracy.

% \subsection{Optimal Parameters}

% We conduct an experiment to evaluate the optimal parameters for each language. Notice that, even though the proposed problem is unsupervised detection of semantic change, our analysis aims to shed light upon the design choices for the model when applied to different languages and different types of semantic change. To that end, we search for the best threshold $t$, number of landmarks $n$, and feature configuration $C$ for each language. The parameter search consists in varying threshold $t$ in $[0.1,0.9]$ in increments of $0.1$, landmarks $n$ in $[300,N]$ in increments of $300$, where $N$ is the size of the intersecting vocabulary and using the features configurations seen in Table \ref{tab:classification}. The optimal parameters are shown in Table \ref{tab:optimal}.

% \begin{table}[t]
% \centering
%     \begin{tabular}{llrrr}
%         Language    &   Features    &   $n$     & $t$   &   Accuracy  \\ \hline % &   Ranking\\ \hline
%          English    &   COS         &   10000   & 0.8   &   0.703     \\  % &   0.292 \\
%          German     &   COS+FREQ    &   2700    & 0.7   &   0.792     \\  % &   0.567\\
%          Latin      &   COS+FREQ    &   900     & 0.2   &   0.675     \\  % &   0.455\\
%          Swedish    &   COS         &   53577   & 0.75  &   0.806    \\ \hline  % &   0.415\\ \hline
%          Overall    &   -           &     -     &    -  &   0.744     \\   %  &   0.432
%     \end{tabular}
% \caption{Optimal model choices for binary classification}
% \label{tab:optimal}
% \end{table}

\section{Conclusions}

We presented a model for unsupervised detection of semantic change based on anomaly detection over a selection of features. \ourmodel~works directly on the input corpora, not requiring language-specific pre-trained models. The model ensemble is agnostic to the feature models, which means any measure of change could be easily incorporated to it, if desired. Our results show that the model parameters must be chosen carefully for each task and language. Particularly, we have shown that the choice of landmarks for alignment is strictly related to the degree of change of a language. In future work, we plan on addressing this issue by developed principled ways of choosing the words to align so that the semantic change is revealed more accurately.

\section{Acknowledgements}
This work was supported by the Rensselaer-IBM AI Research Collaboration (\url{http://airc.rpi.edu}), part of the IBM AI Horizons Network (\url{http://ibm.biz/AIHorizons}).

% {\small {\bf Acknowledgments.}
% This work was supported by the Rensselaer-IBM AI Research Collaboration (\url{http://airc.rpi.edu}), part of the IBM AI Horizons Network (\url{http://ibm.biz/AIHorizons}).}

% The acknowledgments should go immediately before the references.  Do
% not number the acknowledgments section. Do not include this section
% when submitting your paper for review. \\

\bibliographystyle{coling}
\bibliography{ref}

\end{document}